\documentclass[conference]{IEEEtran}
\IEEEoverridecommandlockouts

\usepackage{cite}
\usepackage{amsmath,amssymb,amsfonts}
\usepackage{algorithmic}
\usepackage{graphicx}
\usepackage{textcomp}
\usepackage{multirow}
\usepackage{xcolor}
\newcommand{\methodname}{{\tt{FedCAug}}}
\def\BibTeX{{\rm B\kern-.05em{\sc i\kern-.025em b}\kern-.08em
    T\kern-.1667em\lower.7ex\hbox{E}\kern-.125emX}}

\begin{document}

\title{Federated Out-of-Distribution Generalization: A Causal Augmentation View}
\author{\IEEEauthorblockN{  
        Runhui Zhang\textsuperscript{1},  
        Sijin Zhou\textsuperscript{2},
        Zhuang Qi\textsuperscript{1}\IEEEauthorrefmark{1}
        }
        \IEEEauthorblockA{\textsuperscript{1} School of Software, Shandong University, Jinan, China\\}
        \IEEEauthorblockA{\textsuperscript{2} AIM Lab, Faculty of IT, Monash University, Clayton, VIC, Australia\\}
        Email: zhangrunhuiz@gmail.com, sjzhou1995@gmail.com, z\_qi @mail.sdu.edu.cn
        
\thanks{\IEEEauthorrefmark{1} indicates corresponding author.} 
}

\maketitle

\begin{abstract}
Federated learning aims to collaboratively model by integrating multi-source information to obtain a model that can generalize across all client data. Existing methods often leverage knowledge distillation or data augmentation to mitigate the negative impact of data bias across clients. However, the limited performance of teacher models on out-of-distribution samples and the inherent quality gap between augmented and original data hinder  their effectiveness and they typically fail to leverage the advantages of incorporating rich contextual information. To address these limitations, this paper proposes a \underline{Fed}erated \underline{C}ausal \underline{Aug}mentation method, termed \methodname{}, which employs causality-inspired data augmentation to break the spurious correlation between attributes and categories. Specifically, it designs a causal region localization module to accurately identify and decouple the background and objects in the image, providing rich contextual information for causal data augmentation. Additionally, it designs a causality-inspired data augmentation module that integrates causal features and within-client context to generate counterfactual samples. This significantly enhances data diversity, and the entire process does not require any information sharing between clients, thereby contributing to the protection of data privacy. Extensive experiments conducted on three datasets reveal that \methodname{}{} markedly reduces the model's reliance on background to predict sample labels, achieving superior performance compared to state-of-the-art methods.
\end{abstract}


\section{Introduction}

Federated learning, as an emerging technology, enables collaborative modeling across distributed clients without sharing sensitive data\cite{qi2024attentive,qi2024crosstraining,qi2022clustering,pFedES,FedMRL,FedSSA,FedGH,QSFL,DBLP:conf/cvpr/WangLX0ZZ23,DBLP:conf/iclr/WangXLX0024,DBLP:conf/www/WangJZHR00024,cai2024lgfgad}, thus promoting model training while ensuring data privacy. With the proliferation of edge devices and the growing demand for data-driven insights, federated learning provides an effective solution for leveraging diverse data sources while addressing data security and privacy concerns\cite{gao2022survey,liu2023cross,qi2023cross,li2020federated,qi2025cross,meng2024improving}. However, out-of-distribution (OOD) generalization\cite{wang2022meta,wang2022causal} presents challenges, especially when there are significant data distribution differences between source domains. This can lead to inconsistent optimization directions for local models, making it difficult to aggregate multiple ill-posed learners into a robust model, which negatively impacts its performance in unseen scenarios and limits its generalization ability.


\begin{figure}[t]
\centering
\includegraphics[width=1.0\linewidth]{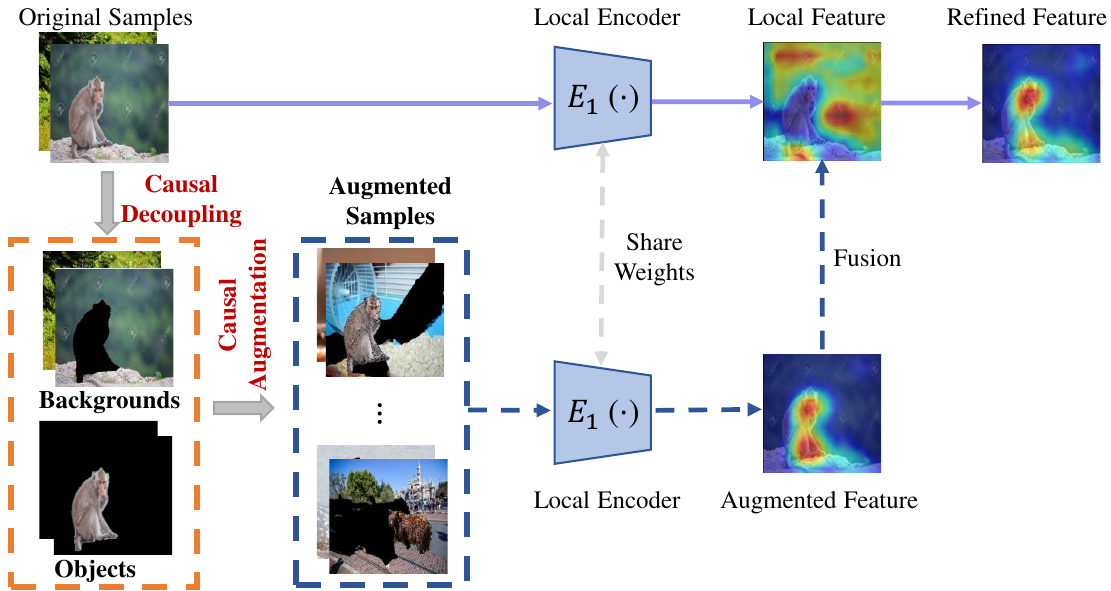}
\caption{\methodname{} can learn better causal representations by reducing background noise interference through causal decoupling and causal augmentation, thereby providing instructive knowledge for the image classification task.} 
\label{fig1}
\vspace{-0.5cm}
\end{figure}

Existing works for addressing the challenges of federated out-of-distribution (OOD) generalization can be broadly categorized into two types: knowledge distillation-based methods and data augmentation-based methods. Knowledge distillation-based methods focus on guiding models to learn domain-invariant features, reducing the influence of irrelevant attributes in image recognition. Techniques such as regularization and feature decoupling are commonly employed in this category. For example, FedCDG\cite{yu2023contrastive} utilizes instance normalization to minimize image differences while enhancing inter-class distinctions and intra-class compactness through prototype contrastive learning. However, this approach may inadvertently weaken the model’s ability to recognize truly domain-invariant features. On the other hand, data augmentation-based methods aim to increase client data diversity, breaking the spurious associations between image noise and labels. These methods often involve generating data samples from other domains by training data generators and exchanging local information. For instance, LADD\cite{shenaj2023learning} proposes a federated clustering aggregation scheme based on client styles to generate diverse data samples, enabling models to learn causal representations in complex scenarios. Despite its potential, this approach raises privacy concerns and is constrained by the quality of the generated data, which can limit its effectiveness.


To address these issues, this study proposes a \underline{Fed}erated \underline{C}ausal \underline{Aug}mentation method, referred to as \methodname{}. Compared to traditional methods, \methodname{} fully leverages client-specific contextual data to enhance sample diversity without the need to share sensitive information between clients. As shown in Figure \ref{fig1}, its core idea is to adaptively decouple the background and objects within an image, establishing connections between each type of background and any category of objects. This approach prevents the model from relying on background information to infer sample labels. Specifically, it designs a causal region localization module to accurately identify and decouple the background and objects in the image. This module provides rich contextual information that serves as the foundation for causal data augmentation, enabling the model to capture meaningful object representations while reducing background interference. The causality-inspired data augmentation module further utilizes these identified causal regions by integrating them into various background images. By generating diverse augmented samples, this module effectively breaks spurious correlations between the background and image labels, ensuring the model relies on causal object representations for prediction, thus improving its robustness and generalization.

Extensive experiments were conducted on three datasets, including performance comparisons, ablation studies of key modules, visual attention case studies of key regions, and error analysis. The results indicate that \methodname{} enhances the model's ability to learn causal representation regions within images, thereby improving overall model performance. In summary, this paper makes two main contributions:
\begin{itemize}

    \item This paper proposes a federated causal augmentation (\methodname{}) method that generates counterfactual samples without sharing any sensitive information between clients. It eliminates spurious associations between background and labels during training, enhancing the model's focus on causal representations.

    \item The proposed \methodname{} is a model-agnostic approach, orthogonal to knowledge distillation-based methods, which can be integrated in a plug-and-play manner into various methods without altering their primary structures. This integration enhances the ability to identify causal representation relationships.
    

\end{itemize}

\section{Related Work}
\subsection{Knowledge Distillation-based Methods}


Knowledge distillation-based methods\cite{huang2023rethinking,luo2022disentangled,zhao2023federated,le2024efficiently} aim to enhance the model's generalization ability by guiding it to learn invariant features across various source domains. By focusing on domain-invariant information, these methods seek to reduce the influence of spurious correlations and enable the model to perform robustly in unseen background distributions. To achieve this, such methods typically employ strategies like regularization or feature decoupling. Regularization-based approaches reduce discrepancies in model outputs across different environments, thereby aligning model predictions with invariant features. Feature-decoupling approaches, on the other hand, aim to separate class-relevant features from irrelevant ones in the feature space, ensuring that only meaningful information is retained for prediction. For example, FedSR \cite{nguyen2022fedsr} introduces L2 norm regularization on representations, which effectively constrains their complexity and prevents the model from capturing spurious correlations arising from domain-specific features. Similarly, DaFKD \cite{wang2023dafkd} adopts a more sophisticated approach by equipping each client with a domain discriminator. This component computes the similarity between samples and their respective domains, allowing the model to perform domain-aware federated distillation. By tailoring knowledge transfer to domain characteristics, DaFKD enhances the model’s ability to adapt to diverse environments.

Despite these methods have demonstrated their potential in improving the generalization of federated models, their effectiveness heavily relies on the availability of domain-specific information. When domain information is sparse or unavailable, the process of invariant feature decoupling becomes less effective, which undermines the model’s ability to recognize and leverage domain-independent features. As a result, the performance gains from these methods may diminish significantly in such scenarios. Therefore, addressing the challenges of limited domain information remains a critical area for improving distillation-based approaches.

\subsection{Data Augmentation-based Methods}

Data augmentation-based methods\cite{park2024stablefdg,chen2023federated,xu2023federated,cubuk2020randaugment} aim to improve model generalization to unseen distributions by increasing the diversity of data attributes and mitigating spurious correlations between attributes and labels. These methods often involve training data generators collaboratively in a federated setting, ensuring the generated samples reflect diverse client styles and underlying data variations. Unlike pre-trained generators, federated training enables synthetic samples to align closely with the distributed client data, promoting broader generalization. Additionally, pre-trained diffusion models are employed to refine existing samples by introducing noise and iteratively enhancing attribute diversity while preserving privacy. For instance, ELCFS \cite{liu2021feddg} adopts a frequency space interpolation mechanism and boundary-oriented learning to share distribution information among clients, generating diverse synthetic samples that improve generalization. Similarly, COPA \cite{wu2021collaborative} optimizes local models for specific domains, aggregates local feature extractors, and integrates domain-specific classifiers, effectively building a unified global model that captures domain-invariant features and mitigates domain-specific biases.

Despite their effectiveness, these methods face challenges such as privacy risks from synthetic data sharing and performance limitations due to the quality gap between generated and real data, requiring better techniques for effective federated data augmentation.


\section{Method}
\subsection{Overall Framework}
\begin{figure*}[t]
\centering
\includegraphics[width=1.0\linewidth]{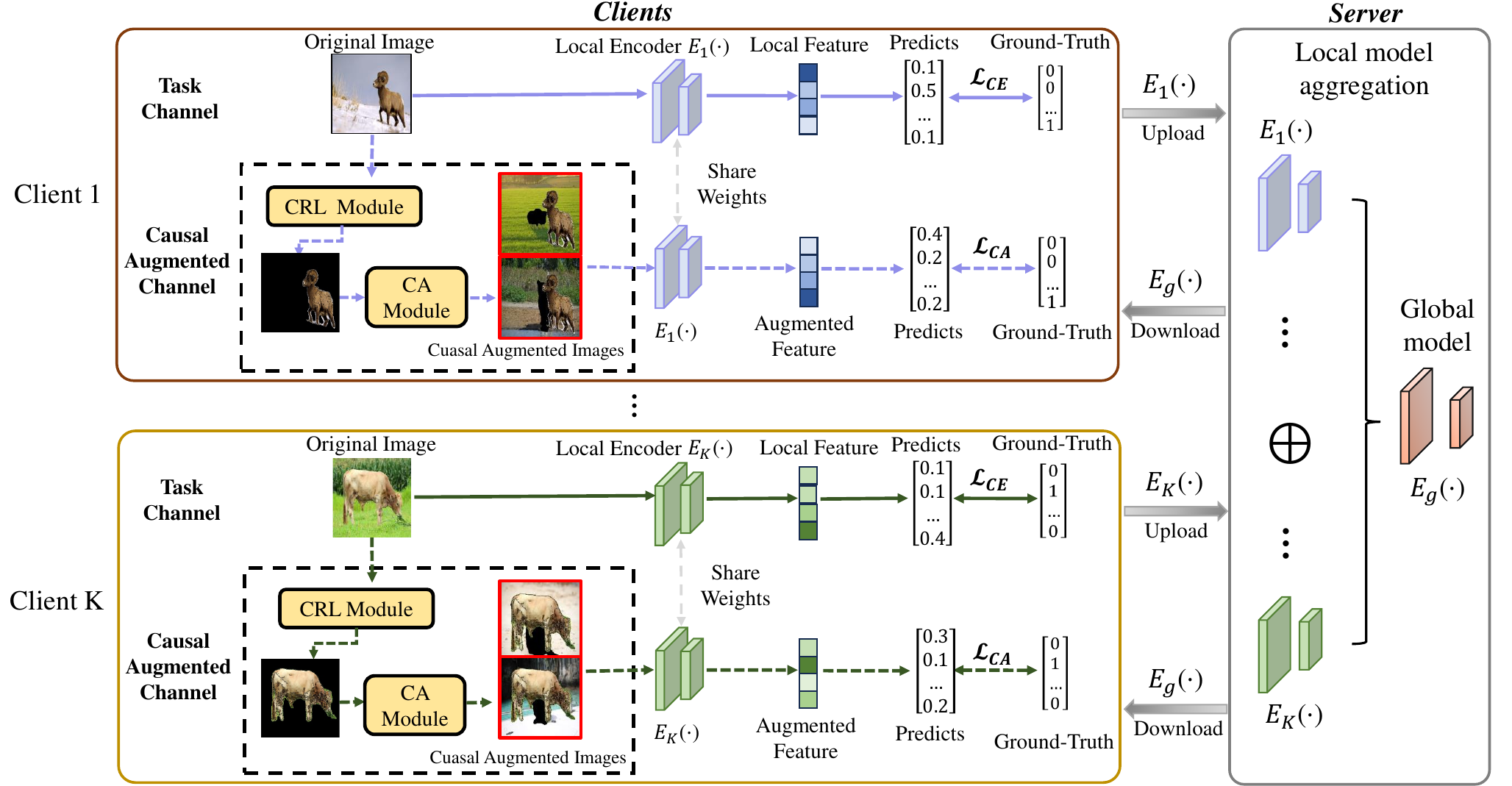}
\caption{Illustration of the framework of \methodname{}, it sharpens images and locates class-relevant regions by using the Causal Region Localization (CRL) module; then it fuses the images with common sense background through the Causal Augmentation (CA) module, providing guiding information for the model to learn causal features.}
\label{fig2}
\end{figure*}

This study proposes a \underline{Fed}erated \underline{C}ausal \underline{Aug}mentation method (\methodname{}), which enhances the model's generalization ability under unseen background distributions. Figure \ref{fig2} illustrates its main framework. Specifically, \methodname{} comprises two key modules: the causal region localization module identifies causal representation regions in images through edge sharpening and saliency detection, effectively refining the boundaries of these regions. Meanwhile, the causal augmentation module integrates the sharpened causal regions into randomly unrelated backgrounds, enhancing the model's robustness and adaptability across diverse unseen scenarios.


\subsection{Causal Region Localization (CRL) Module}

A key challenge in out-of-distribution generalization is the spurious correlation between backgrounds and image labels, which can introduce irrelevant noise during training and degrade the model's performance on unseen distributions. To mitigate this issue, the causal region localization (CRL) module identifies image regions that contain causal representations, directing the model’s attention to these critical areas. Additionally, it enhances the clarity of causal regions, improving the model’s ability to distinguish between the background and the main subjects. The CRL module operates in two stages: image sharpening and causal region segmentation, ensuring more precise localization and refinement of causal features.


Before segmenting the causal regions of the image, the edges of various objects within the image are sharpened through image sharpening stage. This process enhances the clarity of the boundaries between the background and the subject within the image, ensuring that the samples maintain good causal features after being processed by the causal augmentation module while also exhibiting background diversity. Canny edge detection\cite{bao2005canny} is utilized for the image sharpening operation:
\begin{equation}
  I_{edge}= Canny(I),
\end{equation}
\begin{equation}
  I_{sharpened}= I * (1-\lambda_{weighted}) \oplus I_{edge} * \lambda_{weighted},
\end{equation}
where $I_{edge}$ denotes the sharpened edges of the object obtained through Canny detection. $I_{sharpened}$ is represented as the fusion of the original image $I_{O}$ and the sharpened edges $I_{edge}$, $\lambda_{weighted}$  is a weighted parameter.

The causal region segmentation stage decomposes the sharpened sample \(I_{sharpened}\) into object \(I_O\) and background \(I_B\) (\(I_{sharpened} = I_O + I_B\)). Specifically,  a pre-trained PoolNet model \cite{liu2019simple} is used to detect and locate the causal representation regions within the image, thereby isolating the background portions of the image. This process can be expressed as :
\begin{equation}
  I_{CR}= \begin{bmatrix}
  x_1 & x_2 \\
  y_1 & y_2
\end{bmatrix} = PoolNet(I_{sharpened}),
\end{equation}
\begin{equation}
  I_{O}= I_{sharpened} \odot I_{CR} ,
\end{equation}
where $I_{CR}$ is a binary matrix consisting of 0 and 1, 1 is used to represent the identified causal representation region,with $(x_1, y_1)$ as the top-left and $(x_2, y_2)$ as the bottom-right corner.$\odot$ denotes the Hadamard product and $I_O$ denotes the object of the image.


\subsection{Causal Augmentation (CA) Module}
To mitigate the impact of non-independent and identically distributed (non-iid) client datasets and to establish a unified cross-source knowledge base, the causal augmentation module guides the model to learn domain-invariant features by integrating causal representations with background information to generate causal augmented data. 



To achieve this, the object of the image \( I_{{O}} \) is inserted into the random background image \( I_{B}^{random} \) at the sample level. This approach aims to generate causal augmented samples and eliminate spurious associations between the background information and the labels, ensuring that each background information is associated with multiple categories. This process can be expressed as :
\begin{equation}
  I_{CA}= \alpha * I_{O} \oplus (1-\alpha) * I_{B}^{random} ,
\end{equation}
\begin{equation}
  f_{CA}=E_{L}(I_{CA}), f_{I}=E_{L}(I) ,
\end{equation}
where $E_{L}$ denotes the client feature extractor, and $I_{CA}$ represents the causal region augmented image obtained through random background fusion. To simplify the description, the client indices are omitted. $\alpha$ is a hyperparameter, and $f_{CA}$ denotes the features of the causal augmented image, which share the same labels as $f_{I}$. To ensure that the causal augmented images effectively intervene in the model training, we define the causal augmentation classification loss as follows:
\begin{equation}
    \mathcal{L}_{\mathrm{CA}}=-\frac{1}{N} \sum_{i=1}^{N} \sum_{c=1}^{C} y_{c}^{(i)} \log \left(\frac{e^{z_{c}^{(i)}}}{\sum_{j=1}^{C} e^{z_{j}^{(i)}}}\right),
\end{equation}
where $ N $ is the batch size, $ C $ is the number of classes, $ z_c=F_L(f_{CA}) $, $F_L(\cdot)$ is the client classifier, $ y_c^{(i)} $ and $ z_c^{(i)} $ represent the one-hot encoded label and predicted score for class $ c $ of the feature $f_{CA}$, respectively.

Furthermore, to ensure the model's classification capability for the samples, this study employs the standard cross-entropy loss to optimize the client model:
\begin{equation}
  \mathcal{L}_{\mathrm{CE}}=CE(F_{L}(f_{I}),y),
\end{equation}
where  y represents the label of original image I.

\subsection{Training Strategy of \methodname{}}
\methodname{} aims to enhance the model's focus on the subjects within images during training by detecting and augmenting the causal representation regions in the images, thereby improving the model's performance in unseen scenarios. The overall optimization objective can be expressed as:
\begin{equation}
    \mathcal{L}_{total} = E_{(x,y)\sim D_{local}}[\mathcal{L}_{CE} + \mathcal{L}_{CA}]. 
\end{equation}

\begin{table}[t]
    \centering
    \caption{Statistics of COLORMNIST,NICO-Animal and NICO-Vehicle datasets used in experiments,where A7 represents to the data from the first seven backgrounds of each class used as the training set, while B7 represents to the data from the last seven backgrounds of each class used as the training set.}
    \label{tab1}
    \resizebox{0.4\textwidth}{!}{
    \begin{tabular}{c|c|c|c } 
    \hline
     \textbf{Datasets} & \textbf{\#Class} & \textbf{\#Training} & \textbf{\#Testing} \\ 
    \hline
    NICO-Animal (A7) & 10 & 10633 & 2443 \\
    \hline
    NICO-Animal (B7) & 10 & 8311 & 4765 \\
    \hline
    NICO-Vehicle (A7) & 10 & 8027 & 3626 \\
    \hline
    NICO-Vehicle (B7) & 10 & 8352 & 3301 \\
    \hline
    COLORMNIST & 10 & 60000 & 10000 \\
    \hline
  \end{tabular}}\label{tab1}
\end{table}

\section{EXPERIMENTS}
\subsection{Experiment Settings}
\subsubsection{Datasets}
Following the existing work, experiments were conducted on three commonly used datasets: NICO-Animal\cite{NICO_Causal}, NICO-Vehicle\cite{NICO_Causal}, and ColorMNIST~\cite{mnist}. Table \ref{tab1} provides detailed statistics for these datasets.

\subsubsection{Network Architecture}

To ensure a fair and consistent comparison, all methods utilize the same network architecture. In \methodname{}, a unified architecture is employed to process both the original images from the datasets and the causally augmented images. Following prior work~\cite{moon, liu2021feddg}, ResNet-18~\cite{he2016deep} is selected as the backbone network for the NICO-Animal and NICO-Vehicle datasets, while a SimpleCNN, comprising one convolutional layer and two fully connected layers, is used for the ColorMNIST dataset.


\subsubsection{Hyper-parameter Settings}

For all methods in the experiments, the local training epochs per global round were fixed at 10 for the NICO-Animal and NICO-Vehicle datasets, and 5 for ColorMNIST. The number of communication rounds was set to 50, with 7 clients for NICO-Animal and NICO-Vehicle and 5 clients for ColorMNIST. The client sampling fraction was 1.0, and SGD was used as the optimizer. During local training, the batch size was 64, the weight decay was 0.01, and the initial learning rates were set to 0.01 for NICO-Animal and NICO-Vehicle and 0.005 for ColorMNIST. The parameter $\lambda_{weighted}$ was selected from $\{0.1, 0.3, 0.5\}$. The Dirichlet parameter $\beta$ was set to 0.1 and 0.5 for ColorMNIST. For other methods, hyperparameters followed the specifications provided in their respective papers.

\begin{table*}[t]
    \centering
     \caption{Performance comparison between \methodname{} and baseline methods on ColorMNIST, NICO-Animal, and NICO-Vehicle in terms of Top-1 Accuracy. All methods were evaluated over three trials, with both the mean and standard deviation reported.}
     \resizebox{0.85\textwidth}{!}{
     \setlength{\tabcolsep}{3mm}{
    \begin{tabular}{c|c|c|c|c|c|c}
    \hline   
      \multicolumn{1}{c|}{\multirow{2}*{\textbf{Methods}}}& \multicolumn{2}{c|}{\textbf{NICO-Animal}} &
      \multicolumn{2}{c|}
      {\textbf{NICO-Vehicle}}&\multicolumn{2}{c}{\textbf{ColorMNIST}}  \\
     \cline{2-7}
     \multicolumn{1}{c|}{}&A7&B7&A7&B7&\(\beta\)=0.1&\(\beta\)=0.5\\
     \hline
     FedAvg (AISTATS'17)~\cite{fedavg}&44.38±0.6&52.75±0.6 &65.28±0.4&59.05±0.2&89.39±0.6&89.97±0.5\\
     Fedprox (MLSys'20)~\cite{fedprox}&44.55±0.9&51.99±0.9&65.36±0.6&57.50±0.8&86.84±0.4&87.77±0.3\\
     MOON (CVPR'21)~\cite{moon}&45.53±0.4&53.66±0.9&65.94±0.5&59.63±0.5&91.68±0.8&91.15±0.2\\
     FPL (CVPR'23)~\cite{huang2023rethinking}&47.76±0.5&55.39±0.2&68.51±0.7&61.76±0.6&92.94±0.6&\textbf{95.79±0.8}\\
     FedIIR (ICML'23)~\cite{fediir}&46.40±0.9&52.82±0.7&63.64±0.9&56.18±0.4&89.69±0.8&90.23±0.9\\
      FedHeal (CVPR‘24)~\cite{Fedheal}&42.32±1.0&52.80±0.6&64.00±0.5&56.25±0.7&92.02±0.1&87.66±0.6\\
     MCGDM (AAAI'24)~\cite{MCGDM}&47.96±0.8&54.53±0.5&66.84±0.4&59.59±0.9&89.45±0.5&95.40±0.9\\
     
     \hline
     \methodname{}$_\mathrm{FedAvg}$&\textbf{48.47±0.6}&\textbf{55.49±0.2}&\textbf{69.91±0.5}&\textbf{62.46±0.3}&\textbf{93.47±0.5}&95.06±0.8\\
     \hline
     \end{tabular}}}
    \label{tab2}
\end{table*}
\subsection{Performance Comparison}
This section compared the \methodname{} method with seven state-of-the-art (SOTA) methods, including FedAvg\cite{fedavg}, FedProx \cite{fedprox}, MOON\cite{moon}, FPL\cite{huang2023rethinking}, FedIIR\cite{fediir}, FedHeal\cite{Fedheal}, and MCGDM\cite{MCGDM}. Results can be derived from Table \ref{tab2}.

\begin{itemize}[]
\item \methodname{}$_{\mathrm{FedAvg}}$ demonstrates significant improvements in image classification accuracy compared to the baseline method, showcasing its ability to guide the model’s focus toward causal representation regions. This highlights the effectiveness of the proposed causal localization and augmentation mechanisms.
\item The accuracy of \methodname{} in image classification tasks surpasses that of currently existing methods, which is understandable, as \methodname{} enhances the causal representations in images and reduces the negative impact of spurious associations between background information and image labels during training.
\item 
Whether on natural image tasks (NICO-Animal, NICO-Vehicle) or handwritten digit classification (ColorMNIST), \methodname{}$_{\mathrm{FedAvg}}$ demonstrates consistent improvements. Its superior generalization on complex background datasets (NICO) suggests that the causal augmentation strategy effectively mitigates spurious correlations between backgrounds and labels.

\item 
Prototype representations are more effective than global model outputs as regularizers for guiding local model learning as the performance of FPL surpasses that of MOON and FedIIR. This may be due to the poor performance of the global model on out-of-distribution samples.
\end{itemize}

\begin{table}[t]
    \centering
     \caption{Ablation study on the effectiveness of different components of \methodname{} on the NICO-Animal and NICO-Vehicle.}
     \resizebox{0.45\textwidth}{!}{
    \begin{tabular}{c|c|c|c|c}
         \hline
         \multicolumn{1}{c|}{\multirow{2}*{}} & \multicolumn{2}{c|}{\textbf{NICO-Animal}} & \multicolumn{2}{c}{\textbf{NICO-Vehicle}} \\
         \cline{2-5}
         \multicolumn{1}{c|}{}&A7&B7&A7&B7\\
        \hline
         \textbf{FedAvg}&44.38±0.6&52.75±0.6&65.28±0.4&59.05±0.2\\
        \hline

        \textbf{ + $\mathrm{CRL}$ + $\mathrm{CA (CE)}$}&48.47±0.6&55.49±0.2&66.91±0.5&61.46±0.3\\
        \textbf{ + $\mathrm{CRL+CA (CE+Align)}$ }&\textbf{49.86±0.7}&\textbf{55.79±0.5}&\textbf{67.77±0.7}&\textbf{61.58±0.6}\\
         \hline
    \end{tabular}}
    \label{tab3}
\end{table}


\subsection{Ablation Study}
This section further investigates the effectiveness of each module within the \methodname{} framework, with the results summarized in Table \ref{tab3}.
\begin{itemize}[]
\item  Adding Causal Region Localization (CRL) and Causal Augmentation (CA) with Cross-Entropy (CE) significantly improves performance across all settings. This highlights the effectiveness of causal region identification and augmentation in mitigating spurious correlations and improving model generalization.
\item  Incorporating the alignment mechanism (+ Align) further enhances accuracy. For NICO-Animal (A7), accuracy improves from 48.47\% → 49.86\%, and for NICO-Vehicle (A7), from 66.91\% → 67.77\%.
This suggests that aligning representations across different causal augmentations helps the model learn more robust and invariant features, improving generalization.
\item  The improvements are consistent across A7 and B7 settings in both datasets, confirming the robustness of \methodname{}’s components.
Even in B7, where performance is generally higher, adding CRL + CA(CE) + Align still provides additional gains, proving its effectiveness in handling background variations.
%
\end{itemize}

\subsection{Orthogonality of \methodname{} with Knowledge Distillation-based Methods}

Table \ref{tab4} presents the performance of FedAvg, MOON, and FPL after integrating \methodname{}. It is clear that the methods incorporating \methodname{} demonstrate a significant improvement in accuracy for image classification tasks compared to the previous baseline methods. This highlights the effectiveness and orthogonality of \methodname{}, which enhances the model's ability to focus on relevant causal features. Furthermore, we observe that all methods show greater improvements on the NICO-Animal dataset compared to the NICO-Vehicle dataset. This outcome aligns with the findings in Table \ref{tab2}, where the animal class labels are less susceptible to irrelevant background influences. As a result, traditional methods integrated with \methodname{} are better able to leverage the causal augmented samples, facilitating the model's learning of clearer and more accurate causal representations.


\begin{table}[t]
    \centering

     \caption{Comparison of \methodname{} integrated into FedAvg, MOON, and FPL with the original baseline methods.}
     \resizebox{0.45\textwidth}{!}{
    \begin{tabular}{c|c|c|c|c}
         \hline
         \multicolumn{1}{c|}{\multirow{2}*{}} & \multicolumn{2}{c|}{\textbf{NICO-Animal}} & \multicolumn{2}{c}{\textbf{NICO-Vehicle}} \\
         \cline{2-5}
         \multicolumn{1}{c|}{}&A7&B7&A7&B7\\
        \hline
         \textbf{FedAvg}&44.38±0.6&52.75±0.6&65.28±0.4&59.05±0.2\\
        \textbf{\methodname{}$_\mathrm{FedAvg}$ }&\textbf{48.47±0.6}&\textbf{55.49±0.2}&\textbf{66.91±0.5}&\textbf{61.46±0.3}\\
        \hline
        \textbf{MOON}&48.69±0.3&53.66±0.9&65.94±0.5&59.63±0.5\\
\textbf{\methodname{}$_\mathrm{MOON}$}&\textbf{49.47±0.6}&\textbf{57.64±0.7}&\textbf{67.09±0.4}&\textbf{61.32±0.3}\\
         \hline
         \textbf{FPL}&47.76±0.5&55.39±0.2&68.51±0.7&61.76±0.6\\
\textbf{\methodname{}$_\mathrm{FPL}$}&\textbf{49.31±0.8}&\textbf{57.55±0.4}&\textbf{69.63±0.7}&\textbf{62.94±0.6}\\
    \hline
    \end{tabular}}
    \label{tab4}
\end{table}

\begin{figure}[t]
\centering
\includegraphics[width=0.9\linewidth]{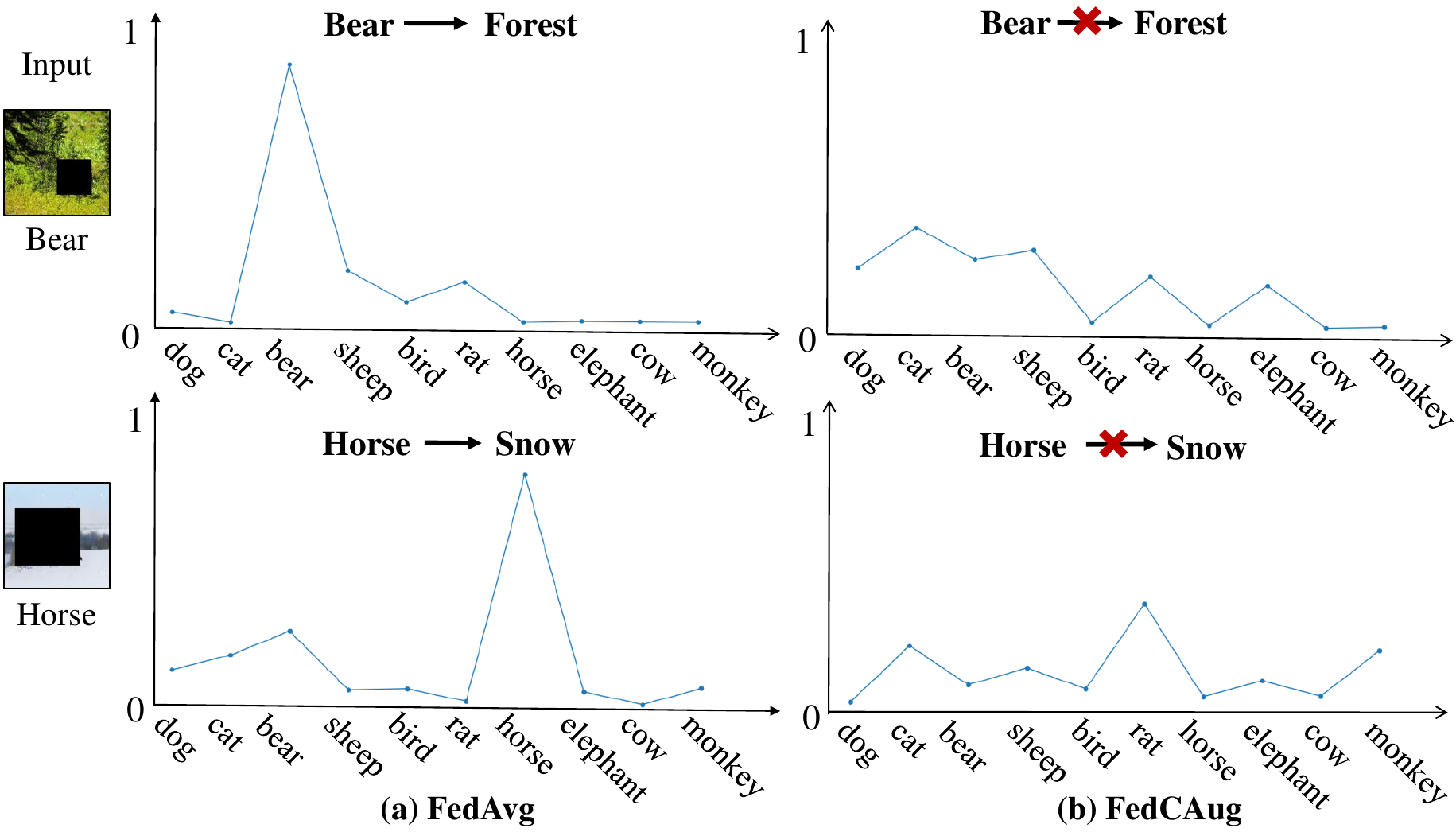}
\caption{The prediction confidence obtained from background images after processing with FedAvg and \methodname{} methods.} 
\label{fig4}
\end{figure}

\begin{figure*}[h]
\centering
\includegraphics[width=0.9\linewidth]{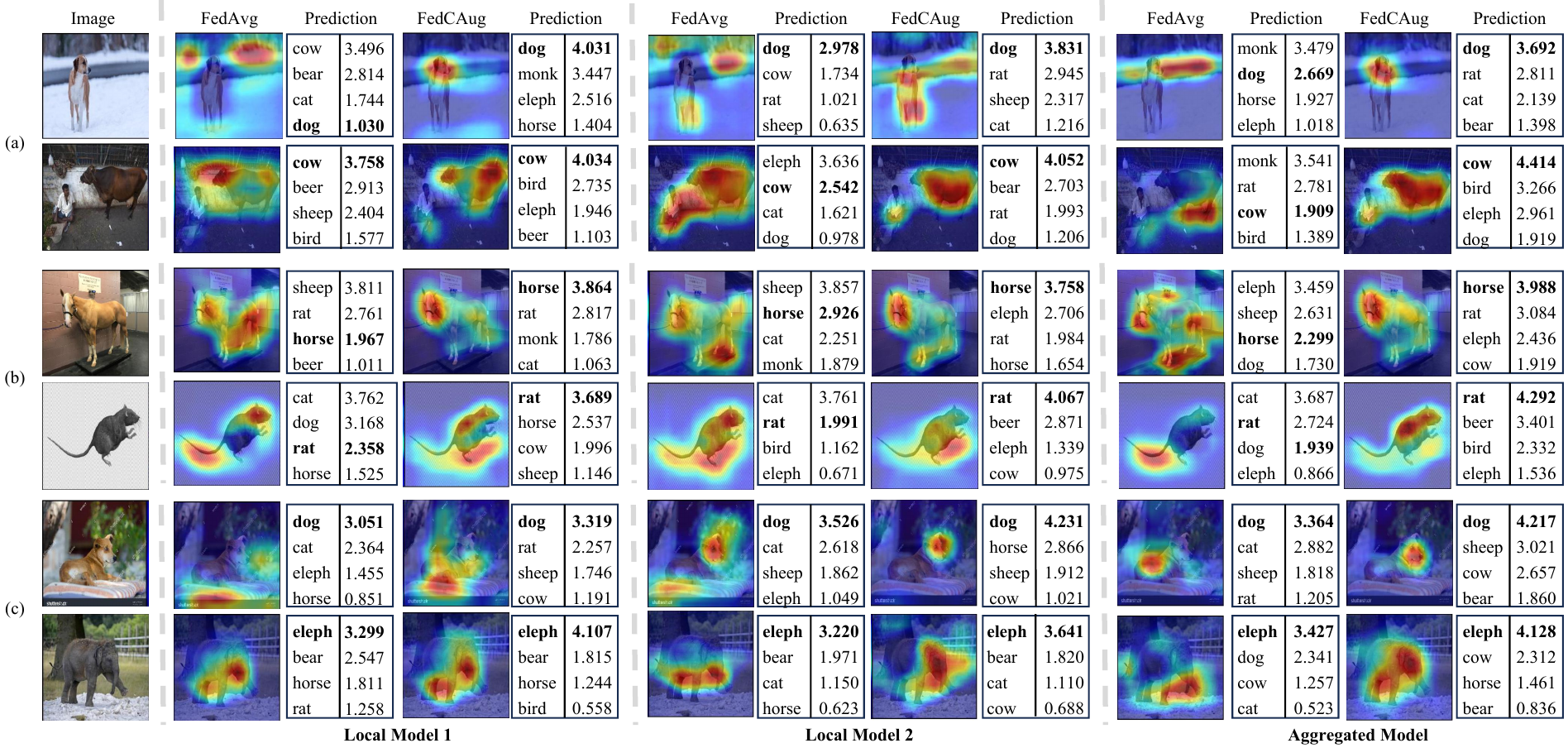}
\caption{Visualization of the visual Attention. (a) The \methodname{} corrects errors in individual clients. (b) The \methodname{} improves the aggregated model by correcting errors in each client, even when both clients make mistakes. (c) The \methodname{} increases the model's confidence in the ground-truth.} 
\label{fig3}
\vspace{-0.5cm}
\end{figure*}

\subsection{Case Study}
\subsubsection{Breaking the Background-Category Association}
This section examines the effectiveness of \methodname{} in mitigating spurious associations between background and labels. To evaluate this, we trained models using FedAvg and \methodname{} and subsequently masked the subjects in the images, retaining only the background as input. The corresponding model predictions are illustrated in Figure \ref{fig4}, revealing that traditional methods tend to overfit to background information, resulting in a strong correlation between the background and image labels. This over-reliance on background cues can lead to incorrect predictions, especially when the model encounters unseen distributions where the background-label correlation no longer holds. In contrast, \methodname{} effectively reduces the model’s confidence in making predictions solely based on the background, encouraging it to focus on learning causal representations. By leveraging causal augmentation, \methodname{} enhances the model’s ability to distinguish between background and essential object features, thereby improving its generalization across diverse environments.


\subsubsection{Visual Attention Visualization}

This section evaluates the effectiveness of the causal augmentation mechanism. Figure \ref{fig3} presents the model outputs of FedAvg and \methodname{}, along with visual attention maps generated using GradCAM \cite{gradcam}. The results clearly demonstrate that \methodname{} enhances the model’s ability to focus on causal representation regions, mitigating the negative influence of background information and correcting prediction errors introduced by FedAvg. In Figure \ref{fig3}(a), the local model trained with FedAvg produces prediction errors, weakening the collaborative learning process and impairing the aggregated model’s decision-making. In contrast, \methodname{} successfully rectifies these errors, improving the overall performance of the aggregated model. Even in Figure \ref{fig3}(b), where both local models trained with FedAvg make incorrect predictions for all images, \methodname{} still effectively corrects these errors, demonstrating its robustness in challenging scenarios. Meanwhile, Figure \ref{fig3}(c) shows that although both FedAvg and \methodname{} correctly classify the image, the model trained with \methodname{} on causally augmented samples exhibits higher prediction confidence, reinforcing the distinction between the ground-truth class and competing categories. These findings highlight the effectiveness of \methodname{} in refining causal representation learning, reducing spurious correlations, and improving the overall predictive performance of both local and aggregated models.


\subsubsection{Showcasing Causal Augmented Samples}

\begin{figure}[h]
\centering
\includegraphics[width=0.98\linewidth]{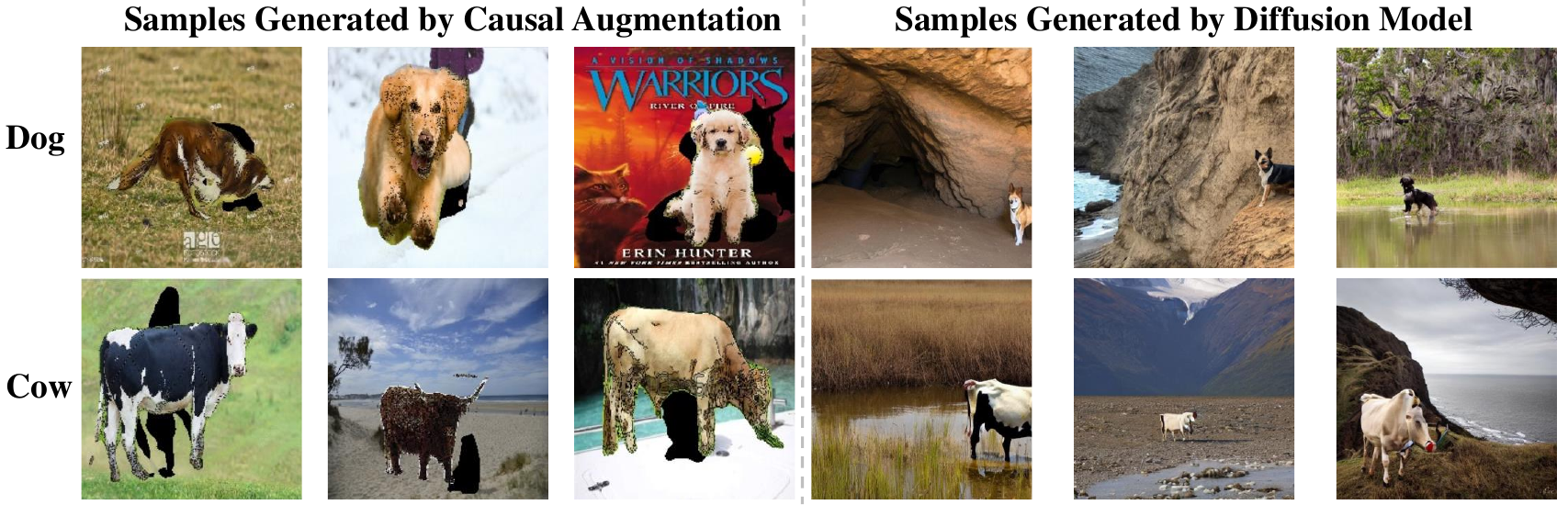}
\caption{A comparative demonstration of causal augmented samples and diffusion model samples.} 
\label{fig5}
\end{figure}


\begin{figure*}[t]
\centering
\includegraphics[width=0.84\linewidth]{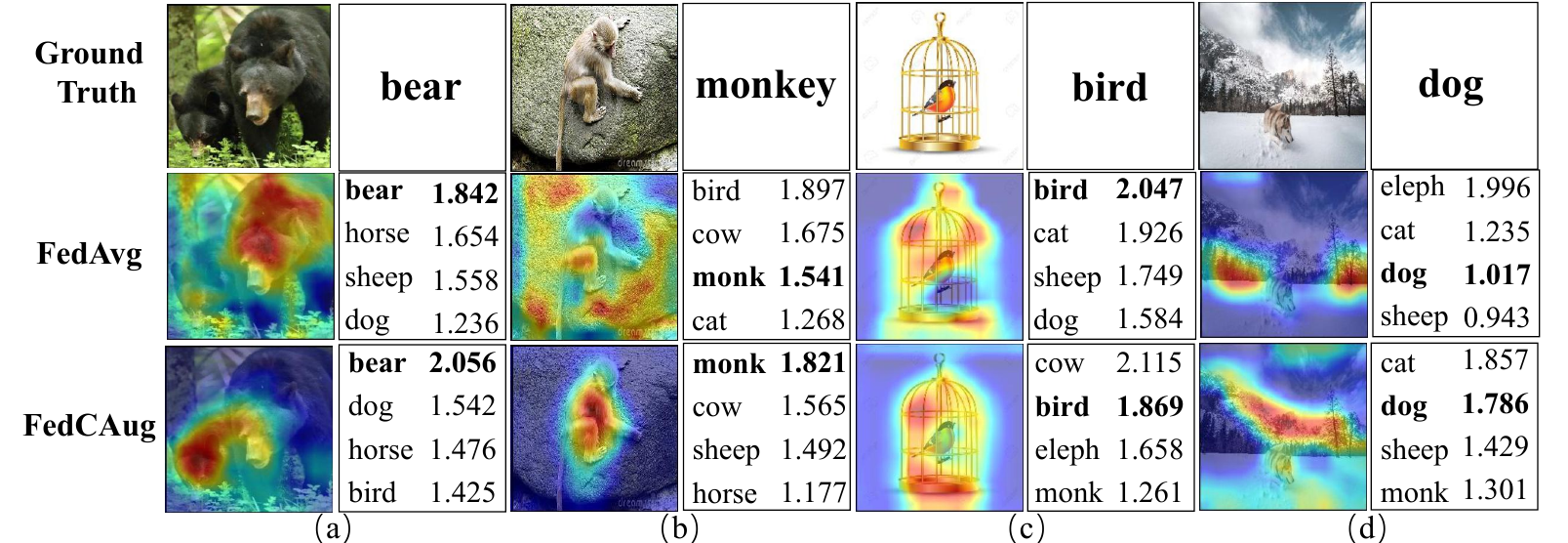}
\caption{Error analysis of \methodname{}. (a) \methodname{} enhances attention to causal regions and increases confidence in predictions. (b) \methodname{} can leverage causal region localization and causal augmentation to correct prediction errors. (c) Smaller visual targets may prevent the model from focusing on the main subject of the image. (d) The presence of multiple subjects in an image can affect the model's ability to focus on the primary subject.} 
\label{fig6}
\vspace{-0.5cm}
\end{figure*}

This section investigates the superiority of causal augmented samples in preserving meaningful causal representations. As illustrated in Figure \ref{fig5}, we randomly select several samples labeled as "dog" and "cow" that have undergone causal augmentation and compare them with images of the same labels generated by a diffusion model. The causal augmented samples exhibit clearer and more distinct causal features of the subjects, effectively retaining the essential characteristics of the objects while minimizing the influence of irrelevant background information. In contrast, the samples generated by the diffusion model often suffer from degraded subject features, where the core object structures become less distinguishable, and the generated images are prone to noise. This results in a diminished ability to capture the underlying causal relationships within the data. The diffusion process, while capable of generating visually diverse images, lacks explicit constraints to enforce the preservation of causal features, leading to inconsistencies in object representation. By leveraging causal augmentation, our approach systematically enhances the focus on subject-relevant attributes while reducing the confounding effects introduced by spurious correlations. As a result, the generated samples not only maintain a more accurate and interpretable causal representation of the subjects, but also contribute to improved model generalization in downstream tasks. These findings underscore the advantages of causal augmentation in mitigating the limitations of generative models that do not inherently preserve causal structures.

\subsection{Error Analysis}

This section analyzes the working mechanism of \methodname{}, focusing on the feature attention derived from Grad-CAM and the model outputs. As shown in Figure \ref{fig6}(a), both FedAvg and \methodname{} make correct predictions, but \methodname{} exhibits a more refined attention on the subjects with clearer causal representations. In Figure \ref{fig6}(b), the FedAvg method is hindered by the complex content of the image, leading to incorrect predictions. In contrast, \methodname{} accurately identifies the causal regions, primarily due to the positive impact of causal-augmented samples in model training. Figure \ref{fig6}(c) reveals that \methodname{} faces challenges with excessive attention to context, making it difficult to distinguish the main subject. However, FedAvg also struggles with focusing on the core object despite making a correct prediction. In Figure \ref{fig6}(d), both FedAvg and \methodname{} encounter difficulties in attending to the complex context of the image; however, \methodname{} is able to reduce its attention to irrelevant areas, effectively narrowing the prediction gap between the "dog" and the top-1 category. This demonstrates the advantages of \methodname{} in image classification tasks and highlights its potential to address the federated OOD problem by promoting clearer causal feature attention and reducing unnecessary noise from complex contexts.


\section{Conclusion}
This paper presents a novel federated learning method based on causal region detection and causal enhancement, termed \methodname{}, to address the federated OOD issue. It conducts representation learning by eliminating image noise, guiding client models to focus on the important regions within the images. Experimental results show that \methodname{} can effectively reduce the interference from background information, and its two modules can be easily integrated. Additionally, this federated learning approach enhances the model's generalization ability in unseen scenarios and the collaborative effect among the models.

Despite the impressive performance of \methodname{}, there are still several directions worth exploring. Firstly, improved causal region detection techniques could more accurately identify the main objects within images. Secondly, introducing imbalanced class data across client models would better reflect real-world scenarios\cite{ma2023cross,li2024cross,wang2023multi,qi2024machine,meng2020heterogeneous,PDRec,TriCDR,li2022unsupervised,wang2024modeling,meng2017towards,li2021comparative,liu2022prompt,qi2024robust}.

\section{Acknowledgments}
This work is supported in part by the Shandong Province Excellent Young Scientists Fund Program (Overseas) (Grant no. 2022HWYQ-048), the Shandong Province Youth Entrepreneurship Technology Support Program for Higher Education Institutions (Grant no. 2022KJN028).

\bibliographystyle{IEEEbib}
\bibliography{reference_related_work.bib}
\end{document}